\RequirePackage{fix-cm}
\documentclass[smallcondensed]{svjour3}     
\smartqed  
\usepackage{graphicx}
\usepackage{amsmath} 
\usepackage[colorlinks,
            linkcolor=blue,
            anchorcolor=blue,
            citecolor=blue
            ]{hyperref}
\usepackage{array}
\usepackage{booktabs}
\usepackage[justification=centering]{caption}
\usepackage{subcaption}
\newcommand{\tabincell}[2]{\begin{tabular}{@{}#1@{}}#2\end{tabular}} 
\begin{document}

\title{Semi-supervised learning method based on predefined evenly-distributed class centroids
}


\author{Qiu-yu Zhu         \and
        Tian-tian Li 
}


\institute{Qiu-yu Zhu \at
              School of Communication \& Information Engineering, ShangHai University, 99 ShangDa Road,BaoShan District, ShangHai City, China \\
              \email{zhuqiuyu@staff.shu.edu.cn}           
           \and
           Tian-tian Li \at
           School of Communication \& Information Engineering, ShangHai University, 99 ShangDa Road,BaoShan District, ShangHai City, China \\
              \email{Twees@shu.edu.cn} 
}

\date{Received: date / Accepted: date}
\maketitle

\begin{abstract}
Compared to supervised learning, semi-supervised learning reduces the dependence of deep learning on a large number of labeled samples. In this work, we use a small number of labeled samples and perform data augmentation on unlabeled samples to achieve image classification. Our method constrains all samples to the predefined evenly-distributed class centroids (PEDCC) by the corresponding loss function. Specifically, the PEDCC-Loss for labeled samples, and the maximum mean discrepancy loss for unlabeled samples are used to make the feature distribution closer to the distribution of PEDCC. Our method ensures that the inter-class distance is large and the intra-class distance is small enough to make the classification boundaries between different classes clearer. Meanwhile, for unlabeled samples, we also use KL divergence to constrain the consistency of the network predictions between unlabeled and augmented samples. Our semi-supervised learning method achieves the state-of-the-art results, with 4000 labeled samples on CIFAR10 and 1000 labeled samples on SVHN, and the accuracy is 95.10\% and 97.58\% respectively. 

\end{abstract}

\section{Introduction}
\label{intro}
In recent years, deep learning has achieved great success in many areas of image research. Deep learning models need to be driven by a large amount of labeled data to achieve good results, for example, using a million-level number of images to train deep neural networks \cite{Ref1,Ref2}. However, collecting labeled samples is difficult for many tasks, which requires a lot of manpower, time, and cost. The remote sensing image classification task can prove this point. In practical applications, the acquisition of labeled data for remote sensing requires not only field investigation but also professional interpretation, which limits the quantity of available labeled samples. In contrast, unlabeled samples are easier to obtain and more numerous, therefore how to use the readily available data to improve the performance of model is an important research issue.

\paragraph{} Semi-supervised learning is a machine learning method between supervised learning and unsupervised learning. In the case of a small number of labeled samples, semi-supervised learning avoids the problem of insufficient model generalization by introducing unlabeled samples. Unlabeled samples provide important information for the spatial distribution of the data, which helps the model to get better decision boundaries. Semi-supervised learning can use only one-tenth or less labeled data to achieve similar results as supervised learning algorithms.

\paragraph{} Recently, semi-supervised learning algorithm is implemented by adding an unlabeled data loss term to the loss function. Pseudo-Label \cite{Ref3} takes the class corresponding to the maximum predicted probability as the true label of the unlabeled sample. However, Pseudo-Label does not use data augmentation, so the results obtained are limited. The earlier semi-supervised methods than Pseudo-Label will not be introduced here, the related overviews are mentioned in literature \cite{Ref4}. In the following, we mainly introduce semi-supervised learning methods with data augmentation.

\paragraph{} Based on the smoothness assumption of the system input and output, a robust model gives a stable and smooth prediction when the input changes (such as shearing, rotation, etc.).  The commonly used regularization method in supervised learning is data augmentation, which obtains lifelike training data by transforming the input without changing the class semantics \cite{Ref5}. Similarly, data augmentation can be applied to unlabeled samples, keeping the output consistentcy before and after augmentation. A teacher-student model inputs noisy samples into the student model, minimizing the prediction error between the teacher model and the student model \cite{Ref6}. Subsequently, the teacher-student model is extended based on the number of iterations to get better results. {$\rm \Pi$}-Model \cite{Ref7,Ref8} updates the prediction of the teacher model by exponential moving average (EMA), and Mean Teacher \cite{Ref9} updates the parameters of teacher model with EMA. However, the data augmentation of {$\rm \Pi$}-Model and Mean Teacher is relatively simple, and only random noise is added to the inputs and hidden layers. Virtual Adversarial Training (VAT) \cite{Ref10} defines the direction of disturbance in the most sensitive direction of the model. MixUp \cite{Ref11} linearly interpolates the inputs and labels of two different samples to obtain augmented samples and labels. MixMatch \cite{Ref12} uses the MixUp method to augment both labeled and unlabeled data, with pseudo-label predicted by the model as the label of unlabeled sample. Nevertheless, VAT and Mixmatch all employ fixed augmentation method for various datasets. UDA \cite{Ref13} is the latest research from google-research, which use AutoAugment \cite{Ref14} to perform data augmentation. AutoAugment is the best-performing data augmentation method. Following UDA, we also use AutoAugment to generate augmented samples. The details of AutoAugment will be introduced in section 2.4. Importantly, the semi-supervised methods introduced above do not add constraints on sample features during network training, resulting in unclear classification boundaries.

\begin{figure}
  \centering
  \includegraphics[width=1.0\textwidth]{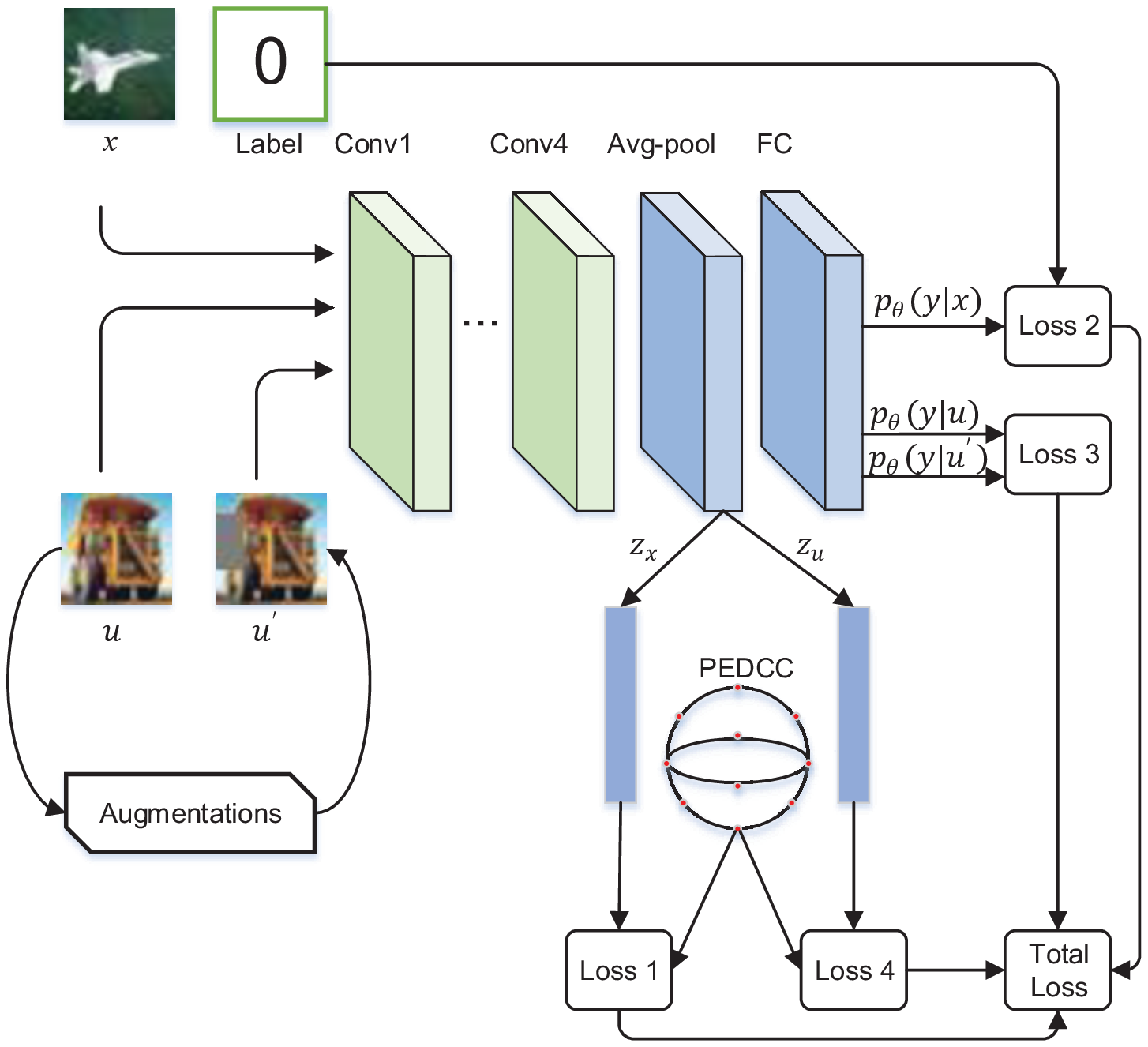}
\caption{Our semi-supervised learning method. The input contains labeled samples, unlabeled samples and augmented samples. The outputs of avg-pool \\ layer are the corresponding feature vectors of the input. PEDCC is the predefined evenly-distributed class centroids}
\label{fig2}       
\end{figure}

\paragraph{} In semi-supervised learning, the loss function consists of two parts: the loss of the labeled samples and the loss of the unlabeled samples. In the field of image classification for supervised learning, in order to get a better feature description, many loss functions have been proposed. The cross-entropy loss function is simple to implement and has good performance, but its feature distribution cannot reach the optimal state. By combining the positive and negative samples, Triplet loss \cite{Ref15} can increase the feature constraints and improve the model generalization, but it takes a long time to train the model. L-Softmax \cite{Ref16} and AM-Softmax \cite{Ref17} change the angle between the weights of the fully connected layer and the features, which makes the intra-class distribution more compact and the inter-class distribution more dispersed. By using predefined evenly-distributed class centroids (PEDCC) and adding the MSE loss between features on AM-Softmax, PEDCC-Loss \cite{Ref18} achieves the best recognition accuracy in CIFAR100, LFW and other datasets. Therefore, we use PEDCC-Loss on a small number of labeled samples and extends its idea to the loss of unlabeled samples. We use the maximum mean discrepancy (MMD) loss \cite{Ref19} to measure the distance between the feature distribution of unlabeled samples extracted by the model and the feature distribution of PEDCC. Therefore, the unlabeled sample feature distribution also satisfies the uniform distribution. Our method makes full use of the labeled samples and unlabeled samples to optimize decision boundaries. The overall diagram of our method is shown in Fig. \ref{fig2}. And the details of the overall diagram are described in section 3.1.

\paragraph{} Our main contributions are as follows:
\paragraph{} 1) The PEDCC is applied to semi-supervised learning. The features of the labeled sample are constrained to the class centroids by the loss function based on the PEDCC-Loss. Our experiments have demonstrated the effectiveness of PEDCC on a small number of labeled samples.
\paragraph{} 2) By adopting AutoAugment data augmentation strategy, the loss function based on MMD is used to restrict the distribution of unlabeled samples to the distribution of PEDCC, and the KL divergence loss function is used to calculate the classification loss between unlabeled samples and augmented samples. The generalization performance of the model is improved by unlabeled samples and augmented samples.
\paragraph{} 3) The conducted experiments show that our semi-supervised learning method achieve the best model prediction accuracy with 4000 labeled samples on CIFAR10 datasets and 1000 labeled samples on SVHN datasets.

\section{Related work}
\label{sec:1}
In this section, we present the work related to the semi-supervised methods we employ.
\subsection{PEDCC}
For deep learning algorithms, the neural network's fitting ability can ensure the intra-class distance is small, but cannot ensure the inter-class distance is large enough. However, if the inter-class distance is small, the accuracy of classification will be reduced. PEDCC is proposed based on the hypersphere charge model \cite{Ref20}. Due to mutual exclusion of charges, in equilibrium, $n$ charges will be evenly distributed on the hypersphere, and the distance between points is the farthest. The purpose of the PEDCC algorithm is to generate $n$ center points that are evenly distributed on the $d$-dimensional hypersphere. $d$ is the feature dimension and $n$ is the number of classification categories. Taking the $n$ center points generated by PEDCC as the clustering center points of $n$ classes can ensure that the inter-class distance is sufficiently large.

\subsection{PEDCC-Loss}
\begin{figure*}
  \begin{subfigure}{.32\textwidth} 
    \centering
    \includegraphics[width=4cm]{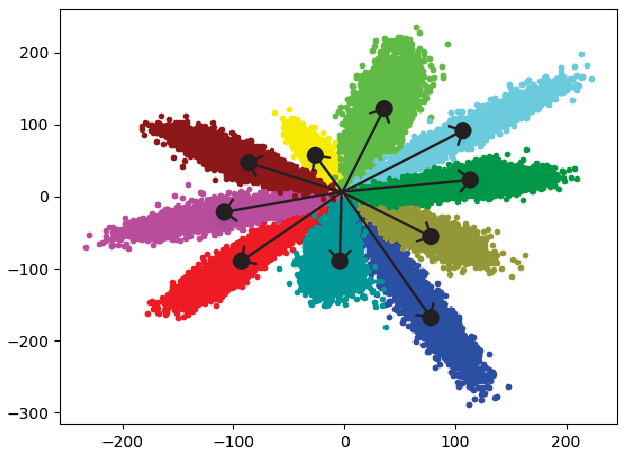}
    \caption{Softmax}
    \label{sf1}
  \end{subfigure}
  \begin{subfigure}{.32\textwidth}
    \centering
    \includegraphics[width=4cm]{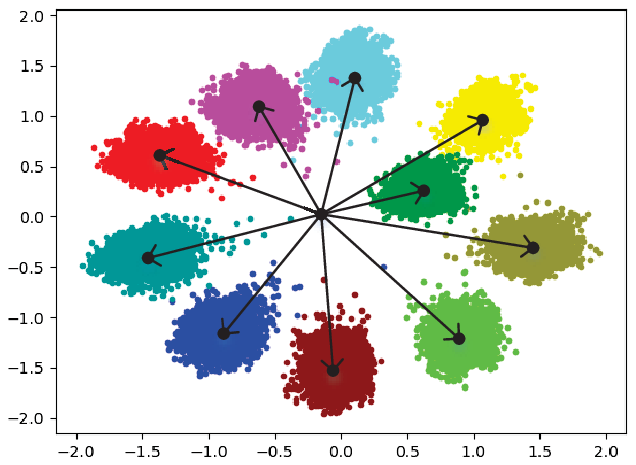}
    \caption{Center Loss}
    \label{sf2}
  \end{subfigure}
  \begin{subfigure}{.32\textwidth}
   \centering
    \includegraphics[width=4cm]{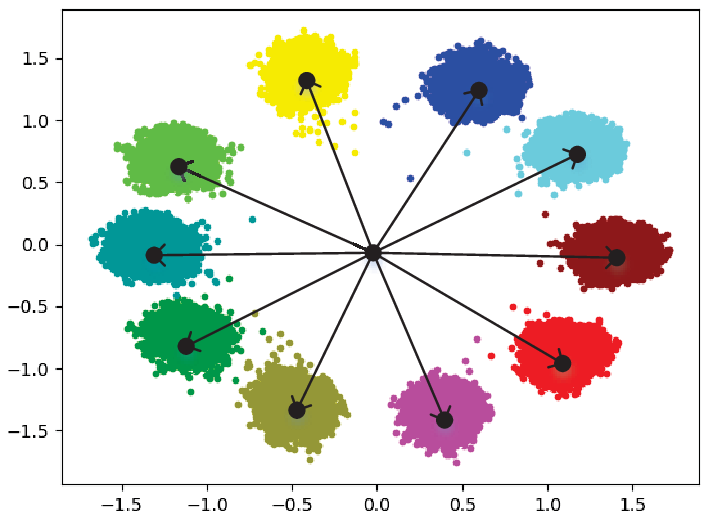}
    \caption{PEDCC-Loss}
    \label{sf3}
  \end{subfigure}
  \caption{The feature distribution finally learned by the neural network using three different loss functions. For ease of visualization, the feature dimension is set to 2}
  \label{fig1}
\end{figure*}

PEDCC-Loss is a loss function based on PEDCC. Fig. \ref{fig1} visualize the feature distribution finally learned by the neural network using different loss functions. It shows that the features learned by PEDCC-Loss have the characteristics of small intra-class spacing and large inter-class spacing. The formulas for PEDCC-Loss are as follows:
\begin{equation}
\label{eq1}
L_{PEDCC-AM}=-\frac{1}{N}\sum_{i} log{\frac{e^{{s\cdot}({\cos\theta_y}_i-m)}}{e^{{s\cdot}({\cos\theta_y}_i-m)}+\begin{matrix} \sum_{j=1,j\ne y_i}^c e^{{s\cdot}{\cos\theta_j}}\end{matrix}}}
\end{equation}

\begin{equation}
\label{eq2}
L_{PEDCC-MSE}=\frac{1}{2}\sum_{i=1}^N{\left \| x_i-{pedcc_y}_i \right \|}^2
\end{equation}
\begin{equation}
\label{eq3}
L=L_{PEDCC-AM}+\lambda\sqrt[n]{L_{PEDCC-MSE}}
\end{equation}
where Eq.(\ref{eq1}) is the AM-Softmax loss function, and $s$ and $m$ are adjustable hyperparameters. Eq.(\ref{eq2}) is the MSE loss function, calculating the distance between the model extraction features and the predefined class features. As shown in Eq.(\ref{eq3}), PEDCC-Loss is obtained by adding the above two loss functions together, and $n$ is a hyperparameter which satisfies $n\ge1$. 

\subsection{Maximum Mean Discrepancy (MMD)}
The mean discrepancy is obtained by finding the continuous function $f$, calculating the mean value of the $f$-mapped samples, and computing the difference between the mean values of the two differently distributed samples. The goal of MMD is to find the function $f$ to maximize the mean discrepancy. As a test statistic, MMD can be used to calculate the distance between two distributions. Therefore. MMD can be used as a tool to determine whether the two distributions are the same. In practical applications, the MMD loss of a batch of data is defined as follows:
\begin{equation}
\label{eq4}
\begin{split}
\begin{aligned}
MMD^2[F,p,q]=&\frac{1}{m(m-1)}\sum_{i\ne j}^m{k(x_i,x_j)}
+\frac{1}{n(n-1)}\sum_{i\ne j}^n{k(y_i,y_j)}\\
&-\frac{2}{mn}\sum_{i,j=1}^{m,n}{k(x_i,y_j)}
\end{aligned}
\end{split}
\end{equation}
\begin{equation}
\label{eq5}
k(x,x')=exp(-\frac{\left \| {x-x'}^2 \right \|}{2\sigma^2})
\end{equation}
where $F$ is the set of all functions $f$, $p$ and $q$ are two different distributions, $x$,$y$ are their corresponding samples. The batch sizes of two distributions are $m$ and $n$, and $\sigma$ is the parameter value of the Gaussian kernel function.

\subsection{AutoAugment}
Using reinforcement learning to search for the best augmentation strategy for a given dataset, AutoAugment improves the effect of manually designed data augmentation method. AutoAugment determines a search space, which consists of multiple sub-policies. Each sub-policy contains two image operations, and each operation has three parts: operation mode, probability, and amplitude. Due to the diversity of image operations, the entire search space has $(16\times11\times10)^{10}$ possibilities. By searching in the search space, AutoAugment finds the most suitable augmentation method for different datasets. For example, the augmentation strategy of CIFAR10 mainly includes color transformation. Meanwhile, the augmentation strategy of SVHN mainly includes geometric transformation. In addition, the augmentation strategy of CIFAR10 can be extended to the strategy of CIFAR100 dataset.

\section{Method}
In this section, we introduce our semi-supervised classification learning method. Our approach incorporates what are presented in the second section.

\subsection{Overall framework}

As shown in Fig. \ref{fig2}, we use labeled samples and unlabeled samples of a given dataset to train the image classification model. Our method first processes the datas{}et, performing data enhancement on the unlabeled sample $u$ to get $u'$. For different datasets, different enhancement strategies are used. The enhancement strategies include a variety of image operations such as rotation, histogram equalization, clipping, and so on. For a given number of classification categories, feature points that are evenly distributed over the hypersphere are generated using PEDCC algorithm.

\paragraph{} The labeled samples $x$, the unlabeled samples $u$ and the enhanced samples $u'$ are input to the convolutional neural network in the same batch according to the specified numbers. The feature description vector of labeled samples $z_x$ and the feature description vector of unlabeled samples $z_u$ are obtained at the output of the pooling layer. The category prediction $p_{\theta}(y|x)$, $p_{\theta}(y|u)$, and $p_{\theta}(y|u')$ are obtained in the final output of the network. Subsequently, the calculation of the loss is performed. For the features of labeled samples $z_x$ and the features of unlabeled sample $z_u$, we add the mean square error loss and the MMD loss respectively, so that the distribution is similar to the PEDCC distribution. Based on the assumption of model-based smoothness, we minimize the KL divergence between the network’s prediction of unlabeled samples and augmented samples. Meanwhile, the AM-Softmax is used to constraint the difference between the predicted values of the labeled samples and the ground truth labels. Through multiple constraints, our algorithm makes full use of the sample to get a clear decision boundary between different classes. The loss function is described in details below.

\subsection{Loss function}
The loss function of our semi-supervised learning algorithm consists of two parts: the loss of the labeled samples and the loss of the unlabeled samples. Each part of the loss is not limited to a single loss.

\subsubsection{Loss of labeled samples}
Usually the loss of the labeled samples is obtained by calculating the error between the output and the ground truth value of the label. However, this constraint can’t ensure the distance of inter-class large enough and the distance of intra-class small enough. We first generate the feature vectors $z_{PEDCC}$ by PEDCC, and its dimension is $C\times D$, where $C$ is the number of classification categories and $D$ is the dimension of the feature vector of each class. Subsequently, the weight of the fully connected layer of the convolutional neural network is fixed to the value of $z_{PEDCC}$, and the weights and features are normalized. Therefore, when the feature vector satisfies a predefined features of a certain class, the output $p_{\theta}(y|x)=wz_x$ of the network is a one-hot vector.

\paragraph{} Assuming that the number of labeled samples in a batch is $M$, we calculate the mean squared error between the feature vectors extracted by the neural network and the predefined centroid as the feature loss function:

\begin{equation}
\label{eq6}
L_1=\frac{1}{M}\sum_{i=1}^M{\left \| {z_x}_i-{z_{pedcc_y}}_i \right \|}^2
\end{equation}
where $i$ is the $i$-th sample in a batch, ${z_{PEDCC}}_y$ is selected from $z_{PEDCC}$ according to the ground truth label $y$, and the feature dimension of ${{z_{PEDCC}}_y}_i$  is $1\times D$. In addition, AM-Softmax is used to calculate the error between the predicted value of the labeled sample and the ground truth label:

\begin{equation}
\label{eq7}
L_2=-\frac{1}{M}\sum_{i} log{\frac{e^{{s\cdot}({\cos\theta_y}_i-m)}}{e^{{s\cdot}({\cos\theta_y}_i-m)}+\begin{matrix} \sum_{j=1,j\ne y_i}^c e^{{s\cdot}{\cos\theta_j}}\end{matrix}}}
\end{equation}
where $\theta$ is the angle between the weights of the fully connected layer and the feature vectors, and $y_i$ is the real label corresponding to the $i$-th sample. $s$ and $m$ are adjustable hyperparameters. The loss of the labeled samples consists of the two losses above:

\begin{equation}
\label{eq8}
L_{labeled}=\lambda_2L_2+\lambda_1\sqrt[n]{L_1}
\end{equation}
where $n$ is an adjustable hyperparameter that satisfies $n\ge1$, and $\lambda_2$ and $\lambda_1$ are used to regulate the relative magnitude between losses.

\subsubsection{Loss of unlabeled samples}
In the supervised learning, the generalization performance of the model is often improved by data augmentation, and the augmented sample shares the same label as the original sample. Similarly, data augmentation can be applied to unlabeled examples of semi-supervised learning. Before and after the augmentation, the predicted output of the unlabeled samples should be consistent. Data augmentation should generate augmented samples that are close to the actual samples, so it is not recommended to use augmented methods that have a large impact on the image, such as adding Gaussian noise. AutoAugment optimizes enhancement strategies for different datasets and learns the most effective enhancement methods for the original datasets. While the previous enhancement methods adopt fixed enhancement strategies for all datasets.

\paragraph{} We use the optimal strategy of AutoAugment to perform data augmentation. Meanwhile, KL divergence is used as a loss function to constrain the distribution consistency between the enhanced samples output and the original samples output:

\begin{equation}
\label{eq9}
L_3=\frac{1}{S}\sum_{i=1}^Sp_{\tilde{\theta}}(y|u_i)log\frac{p_{\tilde{\theta}}(y|u_i)}{p_{\theta}(y|{u_i}')}
\end{equation}
where $S$ is the number of unlabeled samples in a batch. Note that, $S$ is different from the number of labeled samples $M$ in Eq. (\ref{eq6}). $p_{\tilde{\theta}}(y|u_i)$ is the model output of the unlabeled sample, and $p_{\theta}(y|{u_i}')$ is the output of the augmented samples. Following VAT, $\tilde{\theta}$ and $\theta$ share the same value, while $p_{\tilde{\theta}}(y|u_i)$ does not participate in the backpropagation of the model parameters. 
\paragraph{} To further minimize the feature distribution of unlabeled samples and the predefined class centroids, we use the MMD as the loss function:

\begin{equation}
\label{eq10}
\begin{split}
\begin{aligned}
L_4=&\frac{1}{S(S-1)}\sum_{i\ne j}^S{k({z_u}_i,{z_u}_j)}
+\frac{1}{C(C-1)}\sum_{i\ne j}^C{k({z_{PEDCC}}_i,{z_{PEDCC}}_j)}\\
&-\frac{2}{SC}\sum_{i,j=1}^{S,C}{k({z_u}_i,{z_{PEDCC}}_j)}
\end{aligned}
\end{split}
\end{equation}
where the dimension of $z_u$ is $M\times D$ and the dimension of $z_{PEDCC}$ is $C\times D$. $k(\cdot,\cdot)$ is the Gaussian kernel function in Eq. (\ref{eq5}). The loss of the unlabeled samples consists of the two losses above:

\begin{equation}
\label{eq11}
L_{unlabeled}=\lambda_3L_3+\lambda_4L_4
\end{equation}
where $\lambda_3$ and $\lambda_4$ are used to regulate the relative magnitude between losses.

\subsubsection{Final loss}
In our approach, the final loss function of the network is the sum of the labeled samples loss and the unlabeled samples loss:

\begin{equation}
\label{eq12}
L=\lambda_1\sqrt[n]{L_1}+\lambda_2L_2+\lambda_3L_3+\lambda_4L_4
\end{equation}

\subsection{Network structure}
The network structure we use is WideResNet \cite{Ref21}, which reduces the depth and increases the width. Compared to thin and deep ResNet, WideResNet can achieve better image classification accuracy. Table \ref{tab1} lists the specific parameters of the network structure we use. Unlike the previous networks, the weight of fully connected layer is fixed to the predefined evenly-distributed class centroids, and the dimension is $B\times D$. $B$ is the total number of labeled and unlabeled samples in a batch, and $D$ is the feature vectors dimension of each class centroid.

\begin{table}
\caption{Network structure adopted by our algorithm}
\label{tab1}
\centering       
\begin{tabular}{ccc}
\hline\noalign{\smallskip}
Group name&Output size& Block type=B(3,3)  \\
\noalign{\smallskip}\hline\noalign{\smallskip}
Conv1&$32\times32$&$[3\times3,16]$ \\
\specialrule{0em}{1pt}{1pt}
Conv2&$32\times32$&$\left[ \begin{array}{l}3 \times 3,16 \times 2\\3 \times 3,16 \times 2\end{array} \right] \times 4$\\
\specialrule{0em}{2pt}{2pt}
Conv3&$16\times16$&$\left[ \begin{array}{l}3 \times 3,32 \times 2\\3 \times 3,32 \times 2\end{array} \right] \times 4$\\
\specialrule{0em}{2pt}{2pt}
Conv4&$8\times8$&$\left[ \begin{array}{l}3 \times 3,64 \times 2\\3 \times 3,64 \times 2\end{array} \right] \times 4$\\
\specialrule{0em}{1pt}{1pt}
Avg-pool&$1\times1$&$[8\times8]$ \\
\specialrule{0em}{1pt}{1pt}
Fully connected&-&- \\
\noalign{\smallskip}\hline
\end{tabular}
\end{table}

\section{Experiments}

\subsection{Implementation details}
The network structure used in our experiments is WideResNet with depth 28 and width 2. As shown in Table \ref{tab4}, we evaluated our semi-supervised learning algorithms on CIFAR10 \cite{Ref22} and SVHN \cite{Ref23} datasets. Both CIFAR10 and SVHN are benchmark image classification datasets. The total number of sample categories is 10 and the image resolution is $32\times 32$.

\begin{table}
\caption{Datasets details}
\label{tab4}       
\centering 
\begin{tabular}{ccc}
\hline\noalign{\smallskip}
 & CIFAR10 & SVHN  \\
\noalign{\smallskip}\hline\noalign{\smallskip}
Samples &50000  &73257 \\
Labeled samples &4000  &1000\\
Categories  &10  &10 \\
Image Resolution  &$32\times 32$ &$32\times 32$ \\
\noalign{\smallskip}\hline
\end{tabular}
\end{table}

\paragraph{} In the experiment, we used 4000 labeled samples for CIFAR10 and 1000 labeled samples for SVHN. Note that, AutoAugment uses these labeled samples to find the optimal strategy for data enhancement. Therefore, the selection of labeled samples in our experiments is consistent with AutoAugment. For each unlabeled sample, 100 enhanced samples are generated by data augmentation. In one batch, there are 32 labeled samples, 160 unlabeled samples and the corresponding 160 augmented samples. The mode of network learning rate attenuation uses cosine decay. The initial learning rate is set to 0.03 on CIFAR10 and 0.05 on SVHN. The gradient descent method with momentum is used as the optimizer, and the momentum is set to 0.9. All experiments are performed on GTX 1080Ti GPU.

\paragraph{} From Eq. (\ref{eq6}) to Eq. (\ref{eq12}), we can see that there are seven hyperparameters, $s$, $m$, $n$, $\lambda_1$, $\lambda_2$, $\lambda_3$, and $\lambda_4$ involved in our method. Among them, $s$, $m$, $n$, $\lambda_1$, $\lambda_2$ are included in PEDCC-Loss. Therefore, referring to the previous study [16], the value of $s$ is 7.5, the value of $m$ is 0.35. and the values of $n$, $\lambda_1$, and $\lambda_2$ are all 1. Meanwhile, $\lambda_3$ and $\lambda_4$ are used to balance the relative magnitude between different losses. In the following, we will discuss how to determine the values of $\lambda_3$ and $\lambda_4$. The details of the hyperparameter settings are shown in Table \ref{tab5}.

\begin{table}
\caption{The values of hyperparameter settings}
\label{tab5}       
\centering 
\begin{tabular}{cccccccc}
\hline\noalign{\smallskip}
Hyperparameters &$s$ &$m$ &$n$ &$\lambda_1$ &$\lambda_2$ &$\lambda_3$ &$\lambda_4$ \\
\noalign{\smallskip}\hline\noalign{\smallskip}
CIFAR10 &7.5 &0.35  &1 &1 &1 &400 &0.2 \\
SVHN  &7.5 &0.35  &1 &1 &1 &1600  &0.04 \\
\noalign{\smallskip}\hline
\end{tabular}
\end{table}

\subsection{Experimental results}
\subsubsection{Comparison with semi-supervised learning methods}
We compare our method with some representative semi-supervised learning algorithms, using the same number of labeled samples and the same network structure. The average results obtained from three replicate experiments are shown in Table \ref{tab2}. Specifically, we compare our method with five semi-supervised learning methods and the supervised methods. The supervised methods use only labeled samples while a large number of unlabeled samples are not used during training. By comparing our methods with the supervised methods, the improvement is 15.36\% on Cifar10 dataset and 10.41\% on SVHN dataset.

\begin{table}
\caption{The test error rates of the existing algorithm on the CIFAR10 and SVHN data sets. Supervised means that only labeled samples are used.The results of Pseudo-Label, {$\rm \Pi$}-Model, Mean Teacher, VAT are reproduced in the study by Oliver et al \cite{Ref24}}
\label{tab2}       
\centering 
\begin{tabular}{ccc}
\hline\noalign{\smallskip}
 & CIFAR10(4k) & SVHN(1k)  \\
\noalign{\smallskip}\hline\noalign{\smallskip}
Supervised        & 20.26 & 12.83 \\
Pseudo-Label      & 17.78 &7.62 \\
{$\rm \Pi$}-Model & 16.37 & 7.19 \\
Mean Teacher      & 15.87 & 5.65 \\
VAT               & 13.86 & 5.63 \\
UDA &5.34  &3.41 \\
Ours  &\textbf{4.90}  &\textbf{2.42} \\
\noalign{\smallskip}\hline
\end{tabular}
\end{table}
\paragraph{}Pseudo-Label \cite{Ref3}, {$\rm \Pi$}-Model \cite{Ref7}, Mean Teacher \cite{Ref9}, VAT \cite{Ref10} and UDA \cite{Ref13} are representative methods in the semi-supervised field. The details of these methods are described in Introduction. However, the loss function used in those methods can not make the feature distribution learned by network reach an optimal state. Our method use PEDCC, and use PEDCC-Loss for labeled samples, MMD for unlabeled samples to constrains the feature distribution learned by neural network to PEDCC. The dimension of the PEDCC is $10\times 128$ on both datasets. As shown in the table, our method achieves 95.10\% accuracy with 4000 labeled samples on CIFAR10 and 97.58\% accuracy with 1000 labeled samples on the SVHN dataset. Compared with the previous state-of-the-art model UDA, the improvement is 0.44\% and 0.99\% respectively. Our method optimizes the feature distribution of network learning, which improves the accuracy of classification recognition. It is worth mentioning that since the UDA experiment is implemented on google TPU, we run UDA source code three times in our experimental environment and the average results are shown in the table.

\subsubsection{Ablation Study}
In this section, we demonstrate the validity of the loss function we employed on the labeled and unlabeled samples based on the predefined class centroids. We performed ablation experiments on both datasets. The performance of the different loss functions is revealed by adding or removing the corresponding component. Specifically, we have adopted the following combinations of loss functions:
\paragraph{} 1) Cross Entropy on labeled samples and KL divergence on unlabeled samples.
\paragraph{} 2) PEDCC-Loss on labeled samples and KL divergence on unlabeled samples.
\paragraph{} 3) PEDCC-Loss on labeled samples and the sum of KL divergence and MMD on unlabeled samples.
\begin{table}
\caption{Test error rate of different loss functions on CIFAR10 and SVHN datasets}
\label{tab3} 
\centering       
\begin{tabular}{ccc}
\hline\noalign{\smallskip}
 Ablation & CIFAR10(4k) & SVHN(1k)  \\
\noalign{\smallskip}\hline\noalign{\smallskip}
\tabincell{c}{Cross Entropy on labeled and KL divergence \\ on unlabeled} &5.45  &2.98 \\
\tabincell{c}{PEDCC-Loss on labeled and KL divergence \\ on unlabeled}  &4.95  &2.62 \\
\tabincell{c}{PEDCC-Loss on labeled and the sum of \\ KL divergence and MMD on unlabeled} &4.53  &2.48 \\
\noalign{\smallskip}\hline
\end{tabular}
\end{table}
\paragraph{} By comparing the second combination with the first way, the performance improvement brought by PEDCC-Loss is demonstrated. In the same way, by comparing the second combination with the third way, the performance improvement brought by the MMD is revealed. Note that PEDCC-Loss is the sum of the first two terms in Eq. (\ref{eq12}). The results of different combinations are given in Table \ref{tab3}. Obviously, it is effective to predefine the class centroids and apply PEDCC-Loss on a small number of labeled samples.
Compared with the cross-entropy loss, the error rate of PEDCC-Loss decreased from 5.45\% to 4.95\% on the CIFAR10 dataset, and on the SVHN dataset the error rate decreased from 2.98\% to 2.62\%. By adding the MMD constraint, the distribution of the unlabeled samples approaches the uniform distribution, and the error rates on the two datasets can be reduced by 0.42\% and 0.14\% respectively.

\subsubsection{Effect of $\lambda_3$ and $\lambda_4$}
\begin{table}
\caption{Effect of $\lambda_3$ and $\lambda_4$ on CIFAR10}
\label{tab6}       
\centering 
\begin{tabular}{ccc}
\hline\noalign{\smallskip}

$\lambda_3$ &$\lambda_4$ &TER(\%) \\
\noalign{\smallskip}\hline\noalign{\smallskip}
400 &0.1 &4.85 \\
\textbf{400} &\textbf{0.2} &\textbf{4.53} \\
400 &0.4 &5.33 \\
200 &0.2 &5.27 \\
600 &0.2 &5.15 \\
\noalign{\smallskip}\hline
\end{tabular}
\end{table}

\begin{table}
\caption{Effect of $\lambda_3$ and $\lambda_4$ on SVHN}
\label{tab7}       
\centering 
\begin{tabular}{ccc}
\hline\noalign{\smallskip}

$\lambda_3$ &$\lambda_4$ &TER(\%) \\
\noalign{\smallskip}\hline\noalign{\smallskip}
1600 &0.02 &2.48 \\
\textbf{1600} &\textbf{0.04} &\textbf{2.34} \\
1600 &0.08 &2.97 \\
800 &0.04 &2.79 \\
2400 &0.04 &2.84 \\
\noalign{\smallskip}\hline
\end{tabular}
\end{table}
The impact of $\lambda_3$ and $\lambda_4$ on the classification results on CIFAR10 dataset is shown in Table \ref{tab6}. First, we fix the value of $\lambda_3$ to 400 and the values of $\lambda_4$ are set to 0.1, 0.2, and 0.4 respectively. Correspondingly, the test error rates (TER) are 4.85\%, 4.53\%, and 5.33\%. Therefore, the optimal value of $\lambda_4$ is determined as 0.2. Based on this result, we fix the value of $\lambda_4$ to 0.2 and the values of $\lambda_3$ are set to 200, 400, and 600 respectively. Correspondingly, the test error rates are 5.27\%, 4.53\%, and 5.15\%. Obviously, the algorithm obtains the optimal value when the value of $\lambda_3$ is 400 and the value of $\lambda_4$ is 0.2. 
\paragraph{}Similarly, on SVHN dataset, when the value of $\lambda_3$ is 1600 and the value of $\lambda_4$ is 0.04, the algorithm achieves the optimal result, and the test error rate is 2.34\%. The impact of $\lambda_3$ and $\lambda_4$ on SVHN dataset is shown in Table \ref{tab7}.

\section{Conclusions}
In this paper, we apply the PEDCC to semi-supervised learning. Unlike other semi-supervised methods, our method add feature constrains using the corresponding loss functions. Therefore, our method ensures that the inter-class distance is large and the intra-class distance is small to improve the accuracy of classification. Since the final loss function consists of multiple items, we experimentally find the optimal settings for parameters. Additionally, the performance gains of PEDCC on labeled and unlabeled samples are demonstrated separately through ablation experiments. At the same time, our method achieves the state-of-the-art performance on the CIFAR10 and SVHN datasets, using 4000 labeled samples and 1000 labeled samples respectively.

In principle, the effect of data augmentation is directly related to the final performance of semi-supervised learning. The data augmentation strategy we use is AutoAugment, which is currently the best performing data augmentation algorithm. Our method mainly adds feature constraints through different loss functions. However, we have not improved the existing data augmentation strategies. Therefore, the improvement of data augmentation methods will be studied in the future.



%
%



\end{document}